%% file: main.tex
\documentclass[10pt,twocolumn,letterpaper]{article}

\usepackage[pagenumbers]{cvpr} 
\usepackage{esvect}
\usepackage{stfloats}
\usepackage{multirow}
\usepackage{algorithm}
\usepackage[noend]{algpseudocode} 
\usepackage{newfloat}
\usepackage{listings}
\usepackage{pifont}

\newcommand{\rh}[1]{\textcolor{black}{#1}}
\usepackage{threeparttable} 

\input{preamble}

\definecolor{cvprblue}{rgb}{0.21,0.49,0.74}
\usepackage[pagebackref,breaklinks,colorlinks,allcolors=cvprblue]{hyperref}
\def\paperID{30133} 
\def\confName{CVPR}
\def\confYear{2026}

\title{STRNet: Visual Navigation with Spatio-Temporal Representation through Dynamic Graph Aggregation}

\author{
Hao Ren\textsuperscript{1,2}, Zetong Bi\textsuperscript{1}, Yiming Zeng\textsuperscript{1}, Zhaoliang Wan\textsuperscript{2}, Lu Qi\textsuperscript{2,3}\thanks{Corresponding author: Lu Qi (luqi360@whu.edu.cn) and Hui Cheng (chengh9@mail.sysu.edu.cn). This work is supported by the National Key R\&D Program (2022ZD0119602) and the Shenzhen Science and Technology Program under Grant (202504045).}, Hui Cheng\textsuperscript{1}\footnotemark[1]\\
\textsuperscript{1}Sun Yat-sen University \:\:\:
\textsuperscript{2}Insta360 Research \:\:\:
\textsuperscript{3}Wuhan University
}

\begin{document}
\maketitle
\input{sec/00_abstract}    
\input{sec/01_introduction}
\input{sec/02_related}

\input{sec/04_method}
\input{sec/05_experiment}
\input{sec/06_conclusion}
{
    \small
    \bibliographystyle{ieeenat_fullname}
    \bibliography{main}
}
\end{document}

%% file: preamble.tex
\usepackage[accsupp]{axessibility}


%% file: sec/00_abstract.tex
\begin{abstract}
Visual navigation requires the robot to reach a specified goal such as an image, based on a sequence of first-person visual observations. While recent learning-based approaches have made significant progress, they often focus on improving policy heads or decision strategies while relying on simplistic feature encoders and temporal pooling to represent visual input. This leads to the loss of fine-grained spatial and temporal structure, ultimately limiting accurate action prediction and progress estimation. In this paper, we propose a unified spatio-temporal representation framework that enhances visual encoding for robotic navigation. Our approach extracts features from both image sequences and goal observations, and fuses them using the designed spatio-temporal fusion module. This module performs spatial graph reasoning within each frame and models temporal dynamics using a hybrid temporal shift module combined with multi-resolution difference-aware convolution. Experimental results demonstrate that our approach consistently improves navigation performance and offers a generalizable visual backbone for goal-conditioned control. Code is available at \href{https://github.com/hren20/STRNet}{https://github.com/hren20/STRNet}.
\end{abstract}

%% file: sec/01_introduction.tex
\section{Introduction}
\label{sec:intro}
Visual navigation is a core capability for the mobile robot that must operate autonomously in unknown or partially observable environments. 
In tasks ranging from room-to-room indoor navigation~\cite{anderson2018vision, gao2023room}
to open-world outdoor navigation~\cite{shah2021ving, liu2025citywalker},
robots rely on raw image-level observations to make sequential decisions toward visually specified targets. 
This problem setting is central to a wide range of applications, 
such as \rh{embodied exploration}~\cite{malpica2023task, ramakrishnan2021exploration}, autonomous driving~\cite{yasuda2020autonomous, hou2021visual} in intelligent robotics.

Recent advances in visual navigation have emphasized improving decision-making modules, such as goal-conditioned policies~\cite{zhu2017target}, behavior cloning strategies~\cite{shah2023gnm}, or instruction-following frameworks~\cite{long2024instructnav}. 
While these methods have achieved impressive results, they often rely on 
visual encoders originally designed for generic computer vision or video understanding tasks rather than for the rapid, fine-grained control decisions demanded by mobile robots. In practice, these encoders are typically ImageNet-pretrained CNNs followed by simple temporal pooling, which tends to blur crucial geometric and motion cues well before they reach the policy head. The cost of such impoverished representations is measured not in top-1 accuracy but in oscillations, stalls and collisions in the real-world deployment.
Consequently, most efforts have been made to design downstream task heads, leaving the quality and structure of the encoded visual representations insufficiently explored.

\input{figs/teaser}
What makes navigation harder than generic computer vision tasks is the need to reason about irregular spatial layouts and temporally causal events simultaneously. Pooling or average attention smooths away the small optical-flow signals that differentiate \textit{approaching goal} from \textit{moving sideways}, while permutation-invariant self-attention ignores the \textit{topological relations} between doorways, corridors and obstacles(Fig.~\ref{fig:spatial}).
Therefore, they struggle to encode the geometry of scenes and motion in a unified way~\cite{shah2023vint, jia2024mail}.
As shown in Fig.~\ref{fig:teaser}(a), the t-SNE visualization reveals that the feature embeddings generated by the temporal pooling are densely clustered and lack sparsity.
Ideally, the spatial-temporal feature embeddings should exhibit sufficient sparsity to enhance discriminability.

To address this gap, we propose STRNet, a unified spatio‑temporal fusion framework that treats representation learning as a core component of visual navigation. 
Each per‑frame feature, extracted by a shared CNN, is interpreted as a graph where nodes correspond to image regions and edges encode similarity scores that are learned based on visual contrast.
Then, a graph aggregation module can capture fine‑grained spatial geometry. While a temporal fusion module, combining hybrid temporal shift with multi‑resolution contrast, injects motion cues without heavy computation. 
The fused representation drives two lightweight heads: a diffusion‑based policy that synthesizes control actions and a regressor that estimates the temporal distance to the goal.
This unified design supplies downstream decision layers with rich spatial context and precise temporal progression, enabling robust, goal‑conditioned navigation and dependable progress estimation.
The extensive experiments on 2D-3D-S~\cite{armeni2017joint}, Citysim~\cite{koenig2004design}, and GRScenes~\cite{wang2024grutopia} indicate the effectiveness and stability of the representation.
Our contributions are as follows:
\begin{itemize}
    \item We identify the underexplored challenge of weak feature representations in visual navigation, and propose a structured feature fusion framework that is designed to enhance spatial and temporal reasoning prior to the decision-making stage.
    \item We propose a graph-based spatial aggregation module to enhance spatial understanding, and a lightweight temporal fusion module that combines hybrid temporal shift with multi‑resolution contrast, producing a compact yet expressive spatio-temporal representation. 
    \item Through extensive experiments on both simulated and real-world navigation tasks, we demonstrate that STRNet significantly outperforms prior baselines, achieving up to a 70\% higher average success rate compared to NoMaD~\cite{sridhar2024nomad} and establishing the effectiveness and robustness of our proposed representation.
\end{itemize}

%% file: figs/teaser.tex
\begin{figure}[t]
    \centering
    \includegraphics[width=\linewidth]{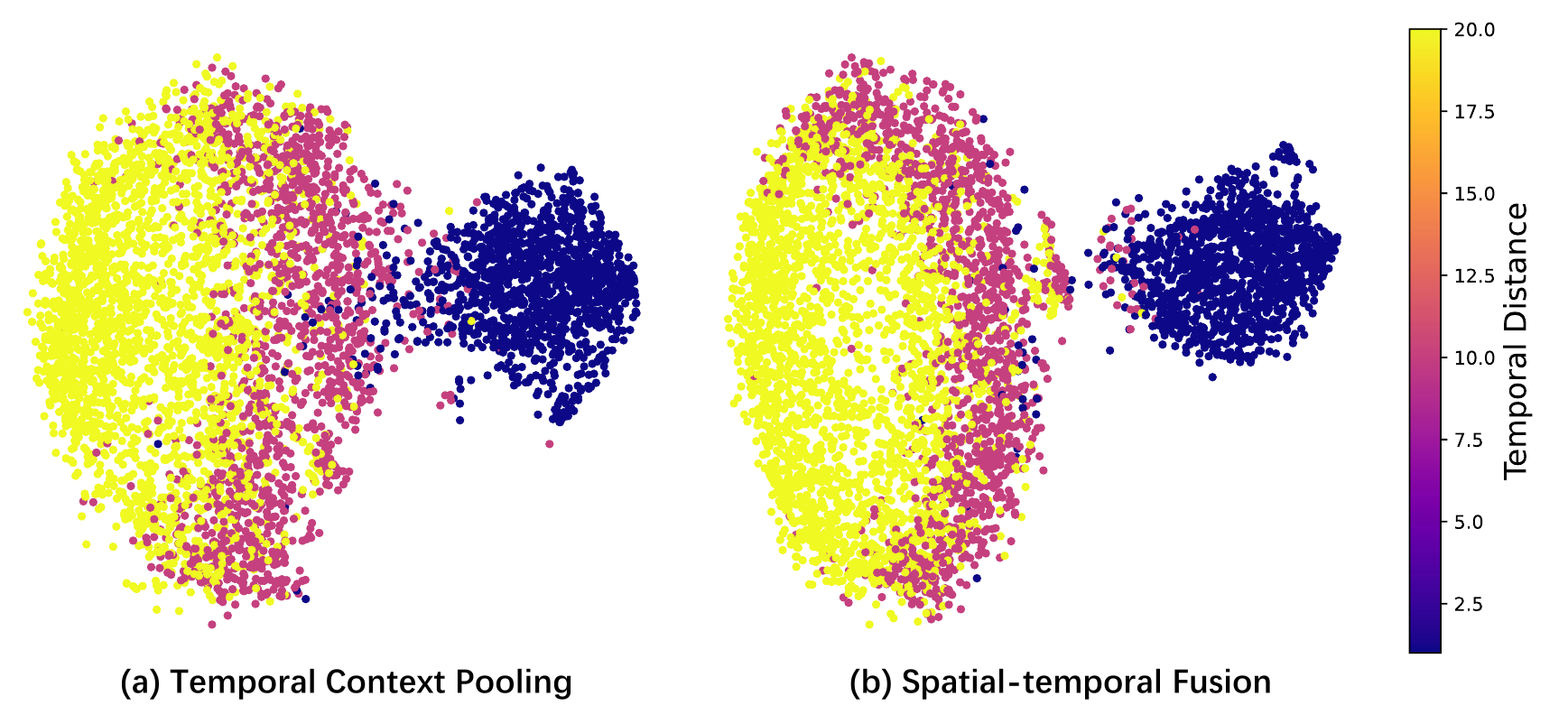}
    \caption{t-SNE projections of feature embeddings colored by ground-truth temporal distances. (a) Conventional temporal-context pooling encoder (NoMaD~\cite{sridhar2024nomad}) produces entangled embeddings, mixing near- and far-to-goal states. (b) Proposed STRNet, using graph-based spatial aggregation and hybrid spatio-temporal fusion, yields clearly separated embeddings, effectively capturing spatial and temporal cues.}
    \label{fig:teaser}
\end{figure}

%% file: sec/02_related.tex
\section{Related Work}
\label{sec:related}

\subsection{Visual Navigation}

Visual navigation is a challenging problem in robotics and artificial intelligence. Traditional methods typically follow a modular pipeline, involving visual mapping, localization, and path planning~\cite{cadena2016past,ren2023adaptive, yang2016survey, yasuda2020autonomous}. While these approaches benefit from well-understood geometric principles, they rely heavily on accurate measurements and often suffer from cumulative error in long-horizon tasks.

Recent learning-based approaches~\cite{zhu2017target, chen2021topological, majumdar2022zson, al2022zero} have enabled end-to-end policies that predict actions directly from visual input, removing the need for explicit mapping or localization. ViNT~\cite{shah2023vint} recently proposed a topological memory model that allows long-range planning and uses diffusion-based subgoal generation, yet still depends on self-attention cascaded dense MLP for feature compression. NoMaD~\cite{sridhar2024nomad} integrates diffusion-based action generation with state-conditioned inputs, achieving strong generalization, yet relies on average pooling for temporal fusion. NaviBridger~\cite{ren2025prior} focuses on the improvement of diffusion policy in action generation, which improves the model performance to some extent. These methods commonly adopt convolutional encoders and use pooling or recurrent networks to aggregate features from observation sequences. Besides, some works improve the performance with multi-modality information (e.g., depth, semantics)~\cite{roth2024viplanner, seymour2021maast, zheng2025get}. However, despite the progress in goal-directed policy learning~\cite{du2021curious, shah2023gnm, zeng2025navidiffusor, wan2025rapid}, the feature encoding stage is often oversimplified, relying on average-pooling or shallow temporal models, resulting in limited capacity to capture rich spatial and temporal cues essential for effective navigation.

In contrast, our method focuses on the quality of visual representation, aiming to preserve and enhance spatial and temporal structure before downstream decision heads.

\subsection{Spatio-temporal Representation Learning}

Effectively modeling spatial and temporal structures in visual data remains challenging. Early methods used recurrent architectures~\cite{mnih2014recurrent} or 3D convolutions~\cite{tran2015learning}, while recent approaches leverage self-attention~\cite{bertasius2021space} and video transformers~\cite{arnab2021vivit} to capture long-range dependencies. However, these models often overlook inherent topological structures. Graph-based methods explicitly model local connectivity and relational reasoning~\cite{myung2024degcn}, benefiting tasks like scene understanding~\cite{yang2018graph} and action recognition~\cite{yan2018spatial}.

\subsection{Representation Learning in Visual Navigation}

Effective representation learning is crucial for interpreting visual inputs in navigation tasks. Recent works include offline visual representation learning via self-supervision~\cite{yadav2023offline}, modeling global environmental context~\cite{li2024memonav}, and aligning latent representations with contrastive learning~\cite{wang2025coal}. Additionally, successor feature representation~\cite{hu2024new}, spatio-temporal region attention mechanisms~\cite{hu2024building}, and spatial attention linking observations, goals, and actions~\cite{mayo2021visual} have significantly improved navigation performance.

%% file: sec/04_method.tex
\section{Methodology}
\label{sec:method}
In this work, we propose a novel visual navigation framework that utilizes a unified spatio-temporal representation for improved visual feature extraction. 
Unlike conventional end-to-end methods, which struggle to fully capture temporal and spatial dynamics, our approach integrates complementary modules to effectively fuse visual information. 
\subsection{Problem Definition}
Given a sequence of past observations \( \mathcal{O} = \{\boldsymbol{I}_t\}_{t=T-p}^{T} \) and a goal image \( \boldsymbol{I}_g \), the goal of visual navigation is to learn a policy \( \pi \) that predicts a control action \( \mathbf{a} \in \mathbb{R}^d \) and a temporal distance estimate \( \tau \in \mathbb{R}_+ \), 
Each observation \( \boldsymbol{I}_t \) is encoded into a feature vector \( \phi(\boldsymbol{I}_t) \), and the resulting sequence is fused into a context representation \( \boldsymbol{c}_T \) that captures both spatial and temporal information. 
The policy first predicts the final action \( \mathbf{a} \) from \( \boldsymbol{c}_T \) by a diffusion policy module. 
The same context is used to estimate \( \tau \), reflecting how close the current state is to the goal in temporal terms.

\input{figs/pipline}
\subsection{Overview of Network Framework}
The architecture of STRNet is shown in Fig.~\ref{fig:pipeline}. 
Given a sequence of RGB observations, STRNet predicts navigation actions and the temporal distances from the target images. 
The framework consists of three stages: visual encoding, spatio-temporal feature fusion, and dual-headed prediction. 
The features of the observation and target images are first extracted individually, and then fused using the proposed spatio-temporal strategy that captures the spatial and temporal relationships essential for effective action planning.
Finally, the fused features $c_T$ feed into the downstream action policy head and temporal distance $\tau$ estimation head. 

\subsection{Spatial Feature Aggregation}
Modern visual backbones typically use convolutional networks (CNNs) or transformers to aggregate spatial features. 
CNNs provide locality and translation equivariance, but have limited receptive fields, restricting long-range interactions. 
Transformers address this limitation through global attention, but bring with high computational complexity and lack structural priors. 
Both paradigms represent visual data as regular grids or sequences, suboptimal for irregular, real-world spatial layouts.

In visual navigation, robots observe semantically meaningful, spatially irregular elements such as doorways, corridors, and obstacles, naturally forming part-whole hierarchies. 
Representing the environment as a graph $\mathcal{G}=(V,E)$—where nodes denote regions and edges denote contextual relationships—captures such structures more effectively, as illustrated in 
Fig.~\ref{fig:spatial}. Vision GNNs thus offer principled, context-aware spatial aggregation~\cite{han2022vision}.

\input{figs/VIG_diagram}
\textbf{Dynamic Axial Graph Construction.}  
Motivated by~\cite{munir2024greedyvig}, we build a dynamic axis-aligned graph that connects nodes only along the horizontal or vertical axes to capture directional relationships within a frame efficiently. 
Given a feature map $X \in \mathbb{R}^{B \times 1 \times A \times A}$, we treat every location $(i,j)$ as a node $x_{i,j}$.
For stride $s$, a circular shift $\mathcal{R}(x_{i,j},s)$ yields a candidate neighbor.
The soft contrast between a pair is as follows:
\begin{equation}
    d_{s}(i,j)=\bigl\lVert x_{i,j}-\mathcal{R}(X,s)_{i,j}\bigr\rVert_{1},
\end{equation}
\begin{equation}
    \label{eq:weight}
    w_{s}(i,j)=\exp\bigl(-d_{s}(i,j)/\tau\bigr),
\end{equation}
where $\tau$ is a temperature. 
Edges are kept implicitly—no hard threshold—by treating $w_{s}$ as soft edge weights.
This produces a sparse, content‑adaptive graph without costly $k$‑NN search.

\textbf{Graph Feature Aggregation.}  
The constructed graph is processed by a hierarchical aggregation block comprising three stages:

\textit{1) Positional Encoding:}  
A depthwise convolution generates a spatial offset map $\phi(X)$, which is added to node features to encode spatial layout:
\begin{equation}
\tilde{X} = X + \phi(X).
\end{equation}

\textit{2) Multi-scale Contrast Enhancement:}  
To further enrich spatial reasoning across varying receptive fields, we introduce a multi-scale directional convolution module. 
For each direction (height or width), we apply circular shifts over multiple strides $s \in \{K, 2K, \dots\}$ and compute contrastive residuals between the shifted and original features. 
For each location, we retain the most salient residual under a contrast-aware masking rule. 
Horizontal directional shifts of multiple strides $s\!\in\!\{K,2K,\dots\}$ are aggregated:
\begin{equation}
    \underbrace{\Delta_{i}}_{\text{horizontal}}
    =\frac{\sum_{s}w_{s}^{i}\cdot\bigl(\mathcal{R}(\tilde{X},s)_{i}-\tilde{X}\bigr)}
           {\sum_{s}w_{s}^{i}+{\varepsilon}},
\end{equation}
where vertical directional shifts are $\Delta_{j}$, the soft residual $\Delta=\max(\Delta_{i},\Delta_{j})$ captures the most salient structural contrasts across scales. 

\textit{3) Residual Transformation:}
The aggregated features are passed through a $1{\times}1$ convolution with normalization and residual connections with the original feature $X$ to increase expressiveness and suppress over-smoothing effects.

Overall, the full spatial aggregation pipeline can be compactly expressed as:
\begin{equation}
\hat{X} = \mathcal{T} \left( \mathcal{A} \left( \mathcal{P}(X) \right) + \tilde{X} \right),
\end{equation}
where $\mathcal{P}(\cdot)$ denotes positional encoding, $\mathcal{A}(\cdot)$ represents contrast-aware graph convolution (including dynamic multi-scale filtering), and $\mathcal{T}(\cdot)$ is a residual transformation block.

This design enables the model to flexibly capture both fine-grained geometry and global context within each frame, yielding structured and content-adaptive features that form a robust foundation for downstream temporal modeling and navigation policy prediction.

\subsection{Temporal Feature Fusion}
While spatial reasoning captures scene structure within individual frames, robust navigation also requires modeling temporal dynamics across a sequence of observations. 
Temporal cues, such as object motion, occlusion transitions, and changes in viewpoint, provide rich context for inferring control signals and estimating goal proximity. 
To this end, we introduce a temporal fusion module that enhances feature representation by combining short-term temporal modeling with multi-scale motion cues aggregation.

Given a sequence of $T$ spatially refined features $\hat{X}_{t}\!\in\!\mathbb{R}^{B\times1\times A\times A}, t=0,1,\dots,T$, Stacking them along the temporal axis yields the 5-D tensor $\mathcal{X} \in \mathbb{R}^{B \times T \times 1 \times A \times A}$.

\textit{1) Hybrid Temporal Shift Module:}
To inject short‑range temporal context with negligible cost, inspired by~\cite{lin2019tsm}, we propose a hybrid temporal shift module that redistributes a small fraction of channels across adjacent frames before a lightweight 3‑D fusion.

Given the per‑frame vector tensor  
$\bar{\mathcal X}\!\in\!\mathbb{R}^{B\times T\times C\times1\times1}$, where $C=A\times A$
we split the channel dimension into four groups  
$C=C_f{+}C_b{+}C_{bi}{+}C_r$ with ratios  
$\{\,\rho,\rho,\rho,\;1-3\rho\}$:
\begin{equation}
    \bar{\mathcal X}
    =\bigl[\,
    \underbrace{\mathcal X^{f}}_{\text{forward}},
    \underbrace{\mathcal X^{b}}_{\text{backward}},
    \underbrace{\mathcal X^{bi}}_{\text{bi‑dir}},
    \underbrace{\mathcal X^{r}}_{\text{residual}}\bigr].
\end{equation}

We perform integer circular shifts along the time axis:
\begin{equation}
    \mathcal X^{f}_{t}= \mathcal X^{f}_{t-1},\quad
    \mathcal X^{b}_{t}= \mathcal X^{b}_{t+1},\quad
    \mathcal X^{bi}_{t}= \tfrac12\bigl(\mathcal X^{bi}_{t-1}+ \mathcal X^{bi}_{t+1}\bigr),
\end{equation}
leaving $\mathcal X^{r}$ unchanged, obtaining shifted features $\widetilde{\mathcal X}$. 
Then a depth‑wise $3\times1\times1$ Conv3D followed by  
a point‑wise $1\times1\times1$ Conv3D fuses the channels,  
and the result is added back:
\begin{equation}
    \mathcal X_{\text{tsm}}
    =\bar{\mathcal X}\;+\;
    \operatorname{Conv}
    \bigl(\operatorname{GN}(\,
    \operatorname{DW\!{-}Conv}(\widetilde{\mathcal X}))\bigr),
\end{equation}
where $\operatorname{GN}$ denotes group normalization and GELU function, producing a motion‑aware tensor  
$\mathcal X_{\text{tsm}}\!\in\!\mathbb{R}^{B\times1\times T\times A\times A}$  
with negligible spatial overhead.

\textit{2) Dynamic Multi-resolution Contrast.}  
To expose motion-related contrasts at different receptive fields, we process $\mathcal X_{\text{tsm}}$ over a pyramid of $K$ spatial scales. 
At scale $k$, we first apply adaptive average-pooling to each frame to obtain a coarser tensor of size $(T,a_k,a_k)$, where $a_k=\max\bigl(A/2^{k-1},4\bigr)$. 
The pooled feature map is then circularly rolled by half its height and width, $(\tfrac{a_k}{2},\tfrac{a_k}{2})$, which shifts foreground structures over background ones and thus amplifies local appearance changes. 
We up-sample the rolled tensor back to $(T,A,A)$ and subtract the original input to form a scale-specific residual:
\begin{equation} 
    \Delta_k\;=\;
    \tilde X_k
    -\mathcal X_{\text{tsm}},
\end{equation}
where$\tilde X_k$ denotes the contrastive counterpart of the original feature after scale-\(k\) receptive-field rearrangement:
\begin{equation}
    \tilde X_k=\mathcal U\bigl(\mathcal R(\mathrm{Pool}_k(\mathcal X_{\text{tsm}}))\bigr).
\end{equation}

All $\{\Delta_k\}_{k=1}^{K}$ tensors are stacked, and a
\emph{cosine-similarity mask} $m_k$ selects spatial locations whose rolled
features are closer than the mean similarity at that scale.

For every space–time index \((t,i,j)\) (the channel dimension is collapsed), we
compute the point-wise cosine similarity
\begin{equation}
\sigma_k(t,i,j)=\mathrm{cos}
\bigl(\tilde X_k(:,t,i,j),\;
      \mathcal X_{\text{tsm}}(:,t,i,j)\bigr),
\end{equation}
and its scale-wise mean:
\begin{equation}
  \bar{\sigma}_k=\frac{1}{T A^{2}}
     \sum_{t,i,j}\sigma_k(t,i,j).
\end{equation}

A hard binary mask keeps the locations whose similarity exceeds the mean,
\begin{equation}
    m_k(t,i,j)=\mathbf 1\!\bigl[\sigma_k(t,i,j)>\bar{\sigma}_k\bigr],
\end{equation}
which is subsequently used to filter the residuals~\(\Delta_k\).


The selected residuals are re-weighted by a learnable scale coefficient
$\beta_k$ and summed:
\begin{equation}
    \mathcal X_{\text{diff}}
    =\sum_{k=1}^{K}\beta_k\,\bigl(m_k\!\odot\!\Delta_k\bigr).
\end{equation}

The resulting \(\mathcal X_{\text{diff}}\) is a multi-scale,
contrast-enhanced feature map that highlights salient temporal changes while
suppressing irrelevant regions.

\textit{3)Contrast‑aware Fusion:}  
Finally, the original and difference tensors are concatenated and fused by a
$1\times1\times1$ Conv3D,
\begin{equation}
    \mathcal X_{\text{fused}}
    =\operatorname{Conv}_{1\times1\times1}\!\bigl(
    [\mathcal X_{\text{tsm}},\mathcal X_{\text{diff}}]\bigr),
\end{equation}
yielding the spatio‑temporal representation supplied to the policy and distance heads.

\subsection{Training Details}
The entire network is jointly trained with supervised learning to predict navigation actions and temporal distances simultaneously. 
The general loss function combines the reconstruction of the action and the estimation of the distance, weighted by a parameter $\alpha$.

For action generation, we use a diffusion policy framework~\cite{chi2023diffusion}, where expert actions are perturbed with Gaussian noise, and the model is trained to iteratively de-noise them. 
The masked mean squared error loss is defined as:
\begin{equation}
\mathcal{L}_{\text{diff}} = \frac{1}{\bar{M}} \sum_i M_i \cdot \|\hat{\epsilon}_i - \epsilon_i\|^2,
\end{equation}
where $M_i$ is a binary mask for valid actions, and $\bar{M}$ normalizes the loss.

For temporal distance estimation, we minimize the mean squared error between predicted and ground-truth distances:
\begin{equation}
\mathcal{L}_{\text{dist}} = \|\hat{d} - d\|^2.
\end{equation}

The combined training objective is:
\begin{equation}
\mathcal{L}_{\text{total}} = \alpha \cdot \mathcal{L}_{\text{dist}} + (1 - \alpha) \cdot \mathcal{L}_{\text{diff}},
\end{equation}
with $\alpha \in [0,1]$ balancing both tasks.

%% file: figs/pipline.tex
\begin{figure*}[htbp]
    \centering
    \includegraphics[width=\linewidth]{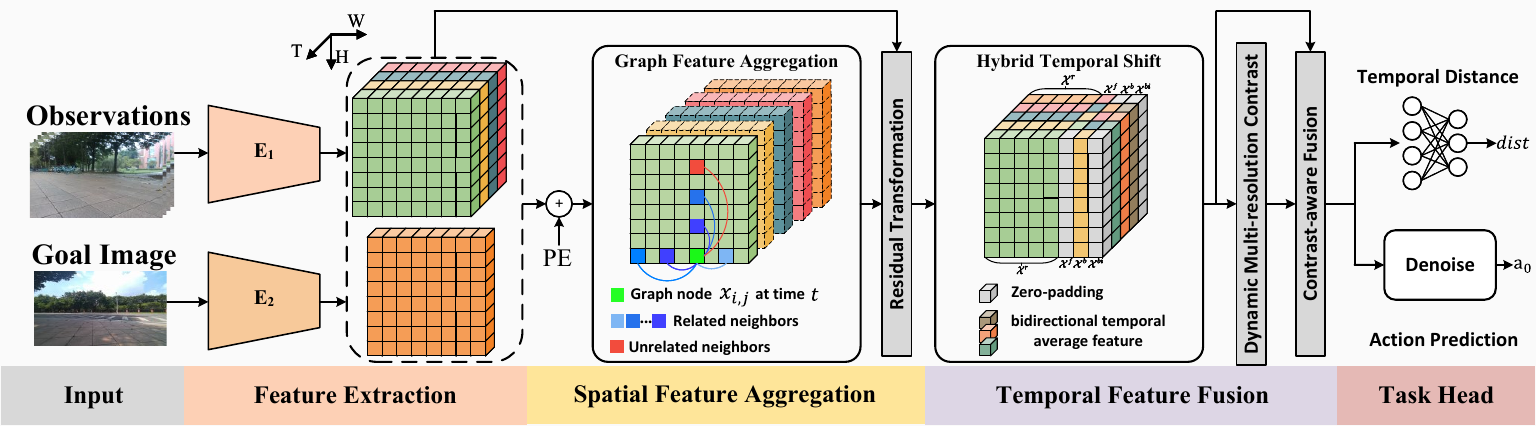}
    \caption{Pipeline of the Proposed Model for Action Prediction: The model processes input observations and goal images through feature extraction, spatial feature aggregation, temporal feature fusion, and hybrid temporal shift, followed by task-specific processing, including temporal distance computation and diffusion denoising to obtain final action prediction.}
    \label{fig:pipeline}
\end{figure*}

%% file: figs/VIG_diagram.tex
\begin{figure}[t]
    \centering
    \includegraphics[width=\linewidth]{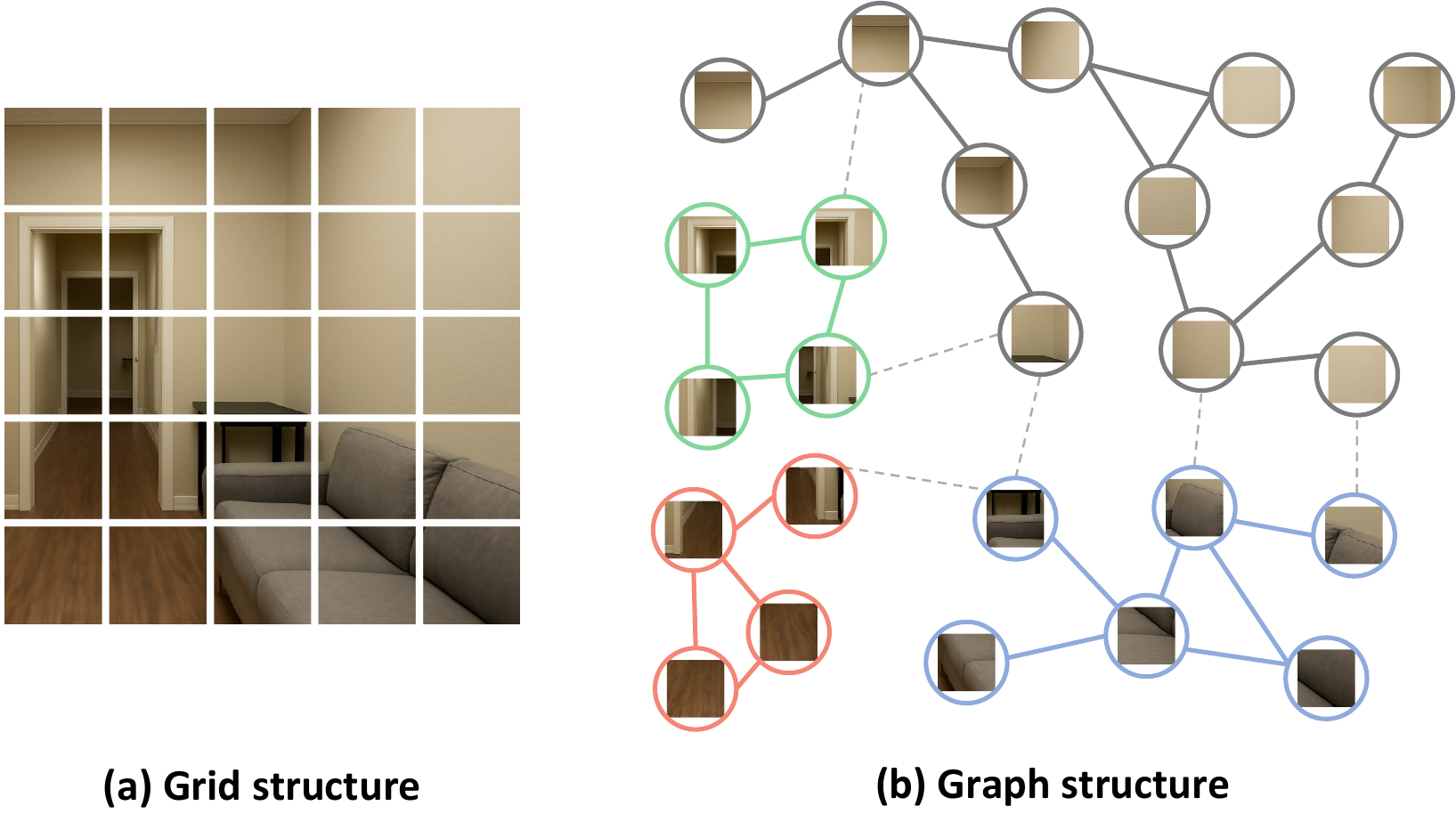}
    \caption{(a) A Grid structure representing a partitioned image, and (b) A Graph structure illustrating the relationships between different regions of the image. The graph structure organizes the context in a more flexible way that aligns with semantic topological relationships, avoiding the local receptive field or predefined order between patches.}
    \label{fig:spatial}
\end{figure}

%% file: sec/05_experiment.tex
\section{Experiments}
\label{sec:experiment}

In this section, we evaluate our method through experiments in indoor and outdoor environments, detailing the experimental setup, results, and ablation studies.

\subsection{Experimental Setup}

\textbf{Datasets}.  
To ensure fair comparison, our method and all baselines were trained on a unified dataset combining data from RECON~\cite{shah2021rapid}, SCAND~\cite{karnan2022scand}, GoStanford~\cite{hirose2019deep}, and SACSoN~\cite{hirose2023sacson}. The dataset includes sequences of image frames with positional information, covering diverse environments and robotic platforms.

\textbf{Baselines}.  
We compare our approach with state-of-the-art methods: ViNT~\cite{shah2023vint}, NoMaD~\cite{sridhar2024nomad}, and NaviBridger~\cite{ren2025prior}. NoMaD integrates diffusion policies with self-attention and average pooling. NaviBridger improves action generation but retains similar feature extraction and fusion strategies as NoMaD. ViNT, a regression-based model, employs self-attention and MLP. Additionally, we evaluate NoMaD-0, a variant that only uses current frame observation image, to assess temporal information utilization.

\textbf{Metrics}.  
We report three key metrics:

\textit{Path Length}: Mean and variance for successful tasks, assessing navigation efficiency and consistency.

\textit{Collision}: Average number of collisions per trial, indicating navigation safety.

\textit{Success Rate}: Percentage of trials where the robot reaches the target within constraints; unsuccessful trials are those with \rh{unresolvable} collisions, timeouts, or missed targets.

\textit{Average SPL}: The average Success weighted by Path Length, which reflects both the success rate and the efficiency of the navigation. Higher SPL indicates that successful trajectories are closer to the shortest possible path.

\textbf{Implementation details}.  
We use the Adam optimizer with cosine annealing, batch size 256, initial learning rate 0.0001, and $\alpha=0.0001$. Input noise is regenerated each epoch, with joint updates to all modules.
The spatio-temporal GNN consists of two temporal-spatial aggregation layers, using multi-scale neighborhoods defined by $K_{\text{list}} = [8, 4, 2]$. The GNN layer count is set to 2 for optimal efficiency and representation capability. Parameters are $\tau=0.1$ for Eq.~\ref{eq:weight} and $\rho=0.125$ for the temporal shift module.

\textbf{Experiment Setuup}.
Comparative experiments were conducted in indoor~\cite{armeni2017joint, wang2024grutopia} and outdoor~\cite{koenig2004design} simulations, as well as real-world tests. Real-world deployment utilized a Diablo robot with NVIDIA Jetson AGX Orin and Azure Kinect (RGB input only). \rh{Every task in the simulation repeats 50 times.}

\subsection{Experiment Results}
\input{figs/environments}
\input{tables/table1}
This section presents an analysis of the experimental results, comparing our method (STRNet) against baselines in various environments. The results include both qualitative and quantitative comparisons, illustrating the strengths and weaknesses of each approach.

\textbf{Qualitative performance}. Figure~\ref{fig:environments} shows qualitative results from both the 2D-3D-S~\cite{armeni2017joint} and Citysim~\cite{koenig2004design} environments. In both cases, we provide the navigation performance of the proposed STRNet model. The \textit{Basic} condition demonstrates the model's ability to navigate short distances, while the \textit{Adaptation} condition shows how the model can generalize to different environments with slight shifts in the conditions. The \textit{Long-range} results (first column, second row) show the model’s ability to maintain stability over longer distances. These results indicate that STRNet performs a stable and efficient navigation in many challenging tasks.

\textbf{Overall quantitative performance}. Table~\ref{table:quantitative} presents a detailed comparison of STRNet with other state-of-the-art methods across two types of tasks: Basic Task and Adaptation Task. The results are presented for two distinct environments: Indoor (2D-3D-S) and Outdoor (Citysim).

The results demonstrate the superior performance of STRNet across both indoor (2D-3D-S) and outdoor (Citysim) environments when compared to baseline methods such as NoMaD, NaviBridger, and ViNT. STRNet's effectiveness stems from its ability to integrate spatio-temporal features, offering a robust model for navigation tasks that require both spatial awareness and temporal context.

Table~\ref{tab:grscenes} demonstrates that STRNet consistently outperforms all baseline methods across every evaluation metric on the GRScenes~\cite{wang2024grutopia} dataset. This dataset consists of high-fidelity indoor simulation scenes. We selected 5 different scenes, and in each scene, two paths were chosen for experimental testing. Each path was repeated 10 times. In terms of Success Rate, STRNet reaches 0.79, which is higher than ViNT and NaviBridger and significantly above NoMaD. This indicates that STRNet is more reliable in completing navigation tasks under the complex spatial layouts of GRScenes.
The average collision metric further highlights STRNet’s advantage. With a value of 0.51, it achieves the lowest collision frequency among all models. Since GRScenes features cluttered scenes and diverse obstacles, this result suggests that STRNet is better at interpreting spatial cues and maintaining stable, safe motion throughout the navigation process.
STRNet also achieves the highest SPL score at 0.80, reflecting more efficient and direct trajectories. The strong SPL performance shows that STRNet not only succeeds more often but does so by following paths that are more coherent and economical, which is important for long range navigation.

\textbf{Comparision with SOTA methods}. In the basic task, STRNet shows consistent improvements across both indoor and outdoor environments. It achieves 100\% success rate in the indoor task, with minimal collision occurrences, outperforming NoMaD and NaviBridger in both path length and collision metrics. Similarly, STRNet excels in the adaptation task, where its performance is notably higher than NoMaD and NaviBridger. Specifically, STRNet reduces collision rates and exhibits superior path efficiency (shorter path lengths) compared to NoMaD. These results underscore the power of STRNet’s spatiotemporal fusion, which allows it to adapt seamlessly to new environments while maintaining optimal navigation paths.

NoMaD's reliance on average pooling for temporal features results in blurred representations and poorer navigation efficiency, particularly in dynamic environments, leading to higher collision rates. Notably, NoMaD-0, without historical observations, often outperforms vanilla NoMaD, underscoring the inadequacy of NoMaD’s temporal fusion strategy.

NaviBridger enhances diffusion-based action prediction but still falls short due to limited temporal feature integration and insufficient spatiotemporal modeling, restricting its adaptability to dynamic scenarios.

ViNT utilizes self-attention and fully connected layers for feature fusion, yet its significantly larger feature dimensions increase computational cost and risk overfitting, reducing effectiveness in adaptation tasks.

In contrast, STRNet effectively integrates spatial and temporal information, delivering superior navigation performance with fewer collisions, shorter paths, and improved adaptability, achieving an optimal balance between efficiency and effectiveness.

\textbf{Long-range Task Analysis}. Table~\ref{table:long} presents the results for the long-horizon task in the 2D-3D-S environment. STRNet excels in this task, outperforming other methods with the shortest length, minimal collisions, and the highest success rate. This further demonstrates the robustness and efficiency of STRNet in complex long distance tasks.

\input{tables/table_isaac}
\input{tables/table_long_horizon}
\input{tables/table_topo_noise}

\textbf{Robustness of sub-target selection}. Stable and effective action prediction significantly impacts local navigation; however, high-level subgoal selection based on temporal distance is equally crucial for long-range navigation performance. To evaluate robustness, we introduce random noise into subgoal selection, creating suboptimal targets. Table~\ref{table:noise} shows that STRNet outperforms other methods under noisy conditions, achieving fewer collisions and higher success rates, highlighting its superior robustness compared to methods like NoMaD, which degrade significantly.

\textbf{Failure case of NoMaD}. Fig.~\ref{fig:compare} visualizes four typical failure modes exhibited by the NoMaD baseline and highlights how an impoverished spatio‑temporal representation can cascade into severe navigation errors. 
(a) Blurred spatial cues fail to distinguish a doorway from adjacent walls, so the agent jitters and stalls at the corner.  
(b) Over‑smoothed temporal features hide loop‑closure evidence, causing the agent to hesitate and wander in circles.  
(c) Weak motion signals prompt over‑steering, yielding an oscillatory, inefficient path.  
(d) Lacking fine detail, the policy overlooks an approaching obstacle and collides with a parked vehicle.  

\subsection{Real-world Evaluation}
\input{figs/compare}
The following experiments show the effectiveness of our method in real-world scenarios using a wheeled-leg robot, Diablo, with an Azure Kinect to obtain images. 
Fig.~\ref{fig:realworld} contrasts the action projections of NoMaD (top row) with those of STRNet (bottom).
Across three representative scenes, (a) a gentle bend, (b) a curb cut, and (c) a cluttered sidewalk, NoMaD issues rigid or hesitant commands, revealing its difficulty in retaining stable and accurate motion signals.
STRNet, empowered by its spatio‑temporal fusion, predicts smooth, goal‑consistent trajectories that align well with the drivable corridor while keeping a safe offset from boundaries.
These qualitative results confirm that richer feature fusion translates into more reliable and human‑like control.

\input{tables/table_ablation}

\subsection{Ablation Studies}
Table~\ref{tab:ablation} confirms the complementary roles of our two design choices.  
When both spatial aggregation (SA) and the proposed temporal fusion block are enabled, the agent finishes the long‑horizon route with a near‑perfect success rate and virtually no collisions.  
Removing either component degrades performance: dropping SA significantly increases the collision rate and lowers success to 88\%, whereas discarding temporal fusion causes severe oscillations that slash success to 38\%.  
When both modules are removed, the policy nearly collapses, achieving only 28 percent success. This extreme degradation clearly illustrates that spatial reasoning and temporal continuity are not interchangeable. Instead, they address two distinct and essential aspects of the navigation problem. Spatial aggregation ensures accurate perception and obstacle interpretation, while temporal fusion maintains coherent decision making over extended sequences. The results collectively show that the strong performance of the full model arises from the complementary effects of these two components. Both are indispensable for achieving safe, smooth, and reliable long-range navigation.

\input{figs/realworld}
\input{tables/table_computational}
\subsection{Computational Analysis}
Table 5 compares the feature extraction model (without task head) complexity and inference efficiency among ViNT, NoMaD, and our proposed STRNet. STRNet achieves the lowest number of parameters compared to ViNT and NoMaD, demonstrating its lightweight design. While STRNet’s FLOPs are slightly higher than NoMaD. The results demonstrate that STRNet achieves a favorable balance between model size, computational cost, and runtime performance. The method retains real-time capability while providing a stronger spatio-temporal encoding than the baselines. This combination of efficiency and representational strength makes STRNet a practical choice for deployment in navigation systems that require both reliability and low latency.

%% file: figs/environments.tex
\begin{figure*}[htbp]
    \centering
    \includegraphics[width=\linewidth]{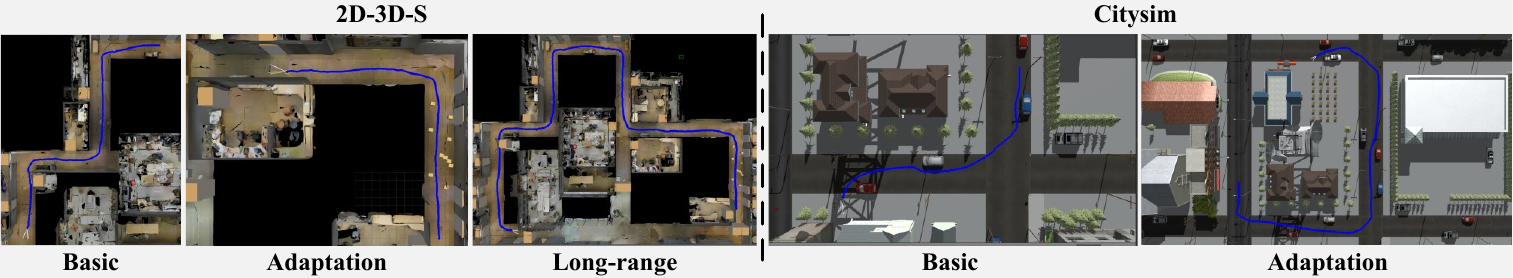}
    \caption{Qualitative navigation trajectories (blue) produced by STRNet in 2D-3D-S and Citysim Environments.}
    \label{fig:environments}
\end{figure*}

%% file: tables/table1.tex
\begin{table*}[ht!]
\centering
\begin{tabular}{cccccccc}
\hline
\multirow{2}{*}{\textbf{Scene}} & \multirow{2}{*}{\textbf{Method}} & \multicolumn{3}{c}{\textbf{Basic Task}} & \multicolumn{3}{c}{\textbf{Adaptation Task}} \\ \cline{3-8} 
& & \textbf{Length (m)} & \textbf{Collision} & \textbf{Success} & \textbf{Length (m)} & \textbf{Collision} & \textbf{Success} \\ \hline

\multirow{5}{*}{\begin{tabular}[c]{@{}c@{}}\textbf{Indoor}\\ (2D-3D-S)\end{tabular}}  
& ViNT & 56.84 $\pm$ 0.098 & \textbf{0} & \textbf{100\%} & 37.60 $\pm$ 0.116 & 0.06 & 80\% \\ 
& NoMaD & 59.08 $\pm$ 2.563 & 0.52 & 70\% & 37.49 $\pm$ 0.278 & 0.38 & 72\% \\ 
& NoMaD-0 & 57.06 $\pm$ 9.384 & 0.24 & 84\% & 39.38 $\pm$ 0.602 & 0.26 & 82\% \\ 
& NaviBridger & 57.60 $\pm$ 1.982 & 0.20 & 90\% & 37.37 $\pm$ \textbf{0.090} & \textbf{0} & \textbf{100\%} \\ 
& \textbf{STRNet (Ours)} & \textbf{56.02} $\pm$ \textbf{0.071} & \textbf{0} & \textbf{100\%} & \textbf{37.07} $\pm$ 0.110 & 0.02 & \textbf{100\%} \\ \hline

\multirow{5}{*}{\begin{tabular}[c]{@{}c@{}}\textbf{Outdoor}\\ (Citysim)\end{tabular}} 
& ViNT & \textbf{49.49} $\pm$ \textbf{0.131} & \textbf{0} & \textbf{100}\% & 192.43 $\pm$ 165.262 & 0.42 & 66\%\\ 
& NoMaD & 58.74 $\pm$ 33.853 & 0.28 & 78\% & 168.82 $\pm$ 128.50 & 1.10 & 36\%\\ 
& NoMaD-0 & 53.31 $\pm$ 0.396 & 0.12 & 92\% & 142.81 $\pm$ 249.32 & 1.14 & 20\%\\ 
& NaviBridger & 59.53 $\pm$ 7.489 & 0.82 & 60\% & 157.16 $\pm$ 97.33 & 0.93 & 42\%\\ 
& \textbf{STRNet (Ours)} & 50.92 $\pm$ 0.770 & \textbf{0} & \textbf{100\%} & \textbf{138.86} $\pm$ \textbf{57.681} & \textbf{0.08} & \textbf{92\%}\\ \hline
\end{tabular}
\caption{Quantitative comparison between the proposed method with baselines in simulation environments}
\label{table:quantitative}
\end{table*}

%% file: tables/table_isaac.tex
\begin{table}[t]
    \centering
    \caption{Comparison of Different Methods on GRScenes}
    \label{tab:comparison}
    \begin{tabular}{lccc}
        \toprule
        Method & \textbf{SR} (\%) & \textbf{Avg. Colli.} & \textbf{Avg. SPL} \\
        \midrule
        VINT & 0.68 & 0.71 & 0.77 \\
        NoMaD & 0.51 & 1.95 & 0.33 \\
        NaviBridger & 0.72 & 0.59 & 0.71\\
        \textbf{STRNet(Ours)} & \textbf{0.79} & \textbf{0.51} & \textbf{0.80} \\
        \bottomrule
    \end{tabular}
    \label{tab:grscenes}
\end{table}

%% file: tables/table_long_horizon.tex
\begin{table}[t!]
\centering
\setlength{\tabcolsep}{2pt}
\caption{Performance metrics for long-horizon task in 2D-3D-S environment~\cite{armeni2017joint}.}
\begin{tabular}{lcccc}
\toprule
\textbf{Method} & \textbf{Length (m)} & \textbf{Collision} & \textbf{Success} \\
\midrule
ViNT                       & 148.53 $\pm$ \textbf{0.420}        & 1.2 & 68\%  \\
NoMaD                      & 159.65 $\pm$ 157.9642   & 1.08 & 30\% \\
NaviBridger & 156.23 $\pm$ 1.164   & 1.73 & 58\%\\
\textbf{STRNet (Ours)}                     & \textbf{145.63} $\pm$ 6.80   & \textbf{0.02} & \textbf{98\%} \\
\bottomrule
\end{tabular}
\label{table:long}
\end{table}

%% file: tables/table_topo_noise.tex
\begin{table}[t!]
\centering
\setlength{\tabcolsep}{4pt}
\caption{Comparison of different methods after adding noise to sub-target selection.}
\begin{tabular}{lcccc}
\toprule
\textbf{Method} & \textbf{Length (m)} & \textbf{Collision} & \textbf{Success} \\
\midrule
ViNT                       & \textbf{19.02} $\pm$ \textbf{0.27}       & 0.52 & 68\%  \\
NoMaD                      & 20.51 $\pm$ 9.29   & 0.90 & 34\% \\
NaviBridger & 19.28 $\pm$ 0.82   & 0.47 & 76\%\\
\textbf{STRNet (Ours)}                     & 19.10 $\pm$ 0.70   & \textbf{0.34} & \textbf{84\%} \\
\bottomrule
\end{tabular}
\label{table:noise}
\end{table}

%% file: figs/compare.tex
\begin{figure}[t]
	\centering
	\subfloat[]{
		\label{fig:subfig:a}
		\includegraphics[height=3cm]{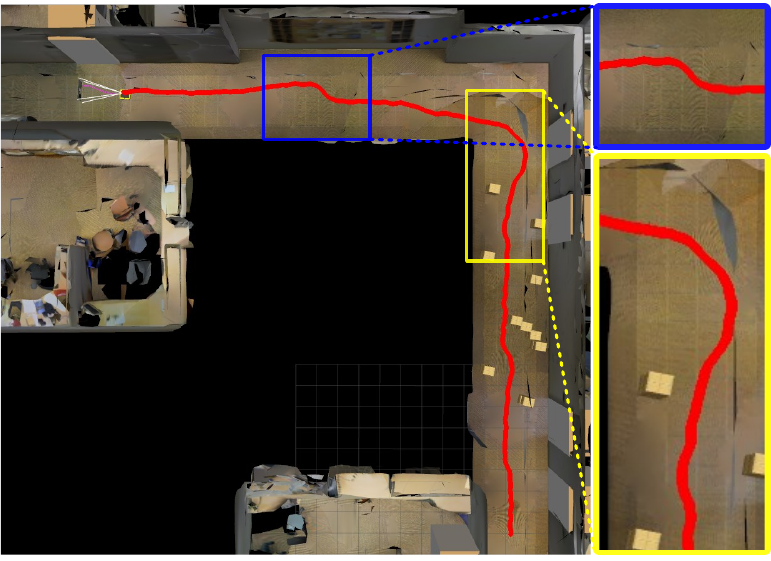}}
	\hspace{0.1in}
	\subfloat[]{
		\label{fig:subfig:b}
		\includegraphics[height=3cm]{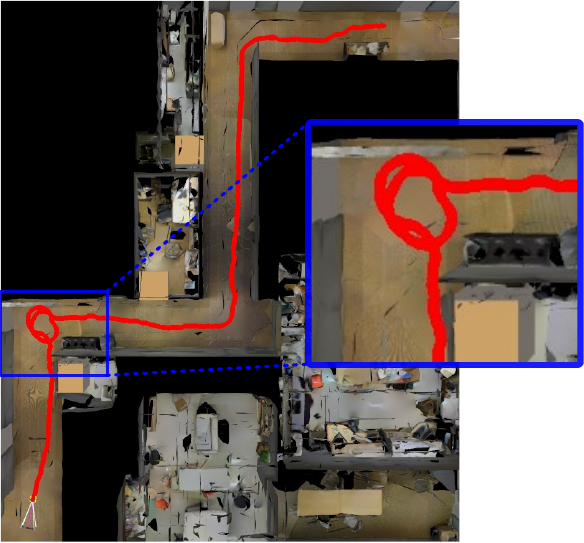}}
	\\
	\subfloat[]{
		\label{fig:subfig:c}
		\includegraphics[height=2.1cm]{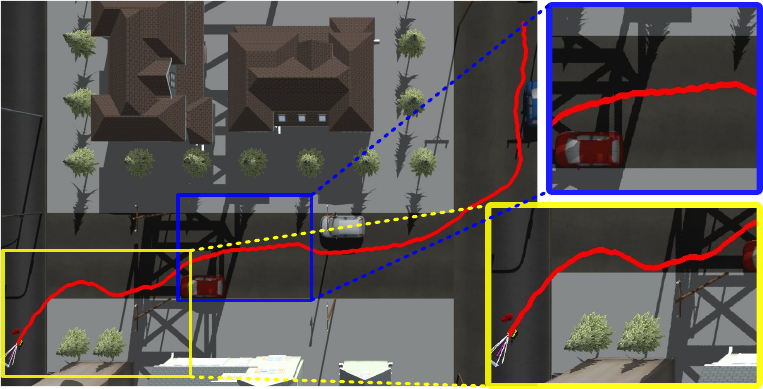}}
	\hspace{0.1in}
	\subfloat[]{
		\label{fig:subfig:d}
		\includegraphics[height=2.1cm]{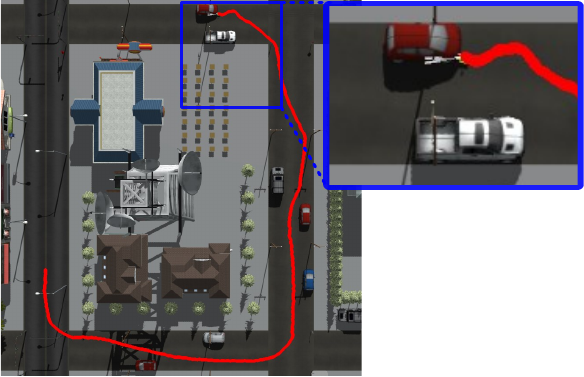}}
	\caption{Failure cases of the NoMaD in visual navigation. (a) Suboptimal behavior caused by poor representation. (b) Hesitation due to incorrect understanding. (c) Incorrect motion direction and erratic trajectory. (d) Increased collisions.}
	\label{fig:compare}
        \vspace{-10pt}
\end{figure}

%% file: tables/table_ablation.tex
\begin{table}[t]
\centering
\caption{Ablation studies on long-horizon task.}
\setlength{\tabcolsep}{2pt}
\begin{tabular}{ccccc}
\toprule
\textbf{SA} & \textbf{Temporal Fusion} & \textbf{Length (m)} & \textbf{Collision} & \textbf{Success} \\
\midrule
 \ding{51} & \ding{51}     & 145.63 $\pm$ 6.80 & 0.02 & 98\% \\
 \ding{55} & \ding{51}     & 144.60 $\pm$ 14.63 & 0.12 & 88\% \\
 \ding{51} & \ding{55}            & 146.28 $\pm$ 4.83 & 0.78 & 38\% \\
 \ding{55} & \ding{55}            & 145.31 $\pm$ 1.15 & 0.86 & 28\% \\
\bottomrule
\end{tabular}
\label{tab:ablation}
\vspace{-10pt}
\end{table}

%% file: figs/realworld.tex
\begin{figure}[tbp]
    \centering
    \includegraphics[width=\linewidth]{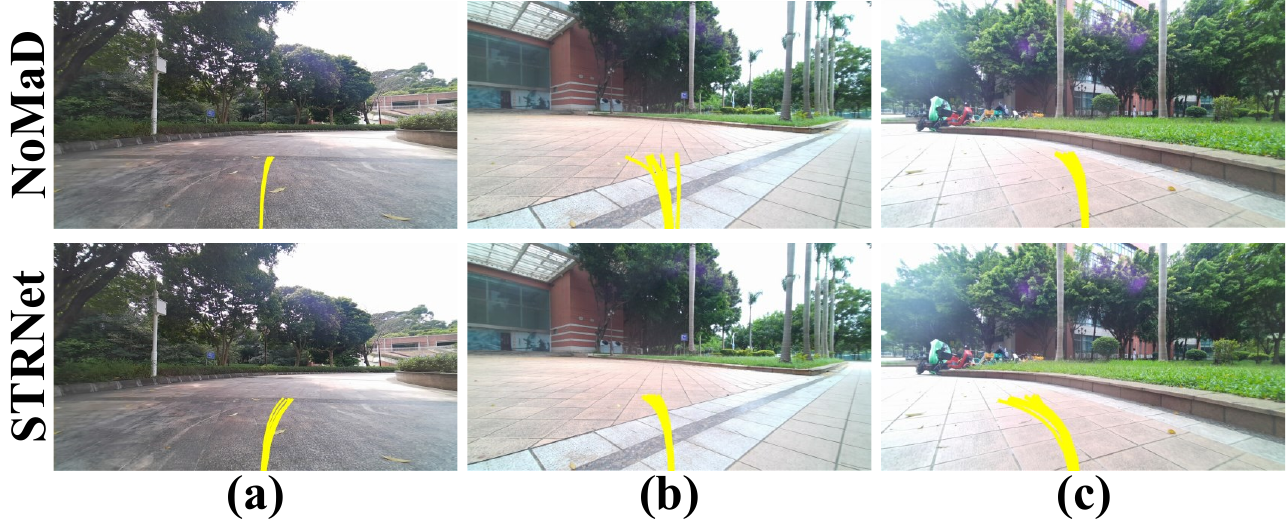}
    \caption{Schematic diagram of front-view projection visualization for action prediction in real-world experiments.}
    \label{fig:realworld}
    \vspace{-10pt}
\end{figure}

%% file: tables/table_computational.tex
\begin{table}[t]
\centering
\renewcommand{\arraystretch}{0.9}
\caption{
Model complexity and inference time comparison.
}
\begin{tabular}{l c c c}
\toprule
\textbf{Method} & \textbf{Parameters} & \textbf{FLOPs} & \textbf{Time} \\
\midrule
ViNT       & 14.04M   & 26.89M         & 30.08ms     \\
NoMaD & 13.60M & 1.70M & 36.06ms   \\
\textbf{STRNet (Ours)}    & 11.83M       & 3.56M       & 36.52ms      \\
\bottomrule
\end{tabular}
\label{table:complexity_comparison}
\vspace{-10pt}
\end{table}

%% file: sec/06_conclusion.tex
\section{Conclusion}
We proposed STRNet, a unified spatio-temporal representation framework that improves goal-conditioned visual navigation by enhancing feature structure across space and time. Unlike prior methods, STRNet introduces a graph-based spatial aggregator and a multi-resolution temporal fusion module, leading to richer visual encodings.

Our dual-headed design supports both action generation and temporal distance prediction. Extensive benchmarks against strong baselines confirm STRNet's consistent gains in efficiency, safety, and success rate. In future work, we plan to extend STRNet to incorporate additional sensory inputs such as depth, lidar, or language commands, and memory mechanisms for more complex navigation tasks.